\documentclass[10pt,journal,compsoc]{IEEEtran}
\usepackage[ruled,vlined]{algorithm2e}
\usepackage{pdflscape}
\usepackage{rotating}
\usepackage{graphicx}
\usepackage{textcomp}
\usepackage{subcaption}
\usepackage{tablefootnote}
\usepackage{threeparttable}
\usepackage{multirow}
\usepackage{etoolbox}
\usepackage{url} 
\usepackage{makecell}
\usepackage{authblk}
\usepackage{xcolor}

\ifCLASSOPTIONcompsoc
  \usepackage[nocompress]{cite}
\else
  \usepackage{cite}
\fi
\ifCLASSINFOpdf
\else
\fi
\begin{document}
%
\title{MOHAQ: Multi-Objective Hardware-Aware Quantization of Recurrent Neural Networks}
%
%
%
%

\author{\uppercase{Nesma M. Rezk}\IEEEauthorrefmark{1}, 
\uppercase{Tomas Nordström\IEEEauthorrefmark{2},Dimitrios Stathis\IEEEauthorrefmark{3},Zain Ul-Abdin\IEEEauthorrefmark{1},Eren Erdal Aksoy\IEEEauthorrefmark{1}, and Ahmed Hemani\IEEEauthorrefmark{3}}  
\\\
\IEEEauthorrefmark{1}Halmstad University, Sweden, \{nesma.rezk,zain-ul-abdin,eren.aksoy\}@hh.se  

\IEEEauthorrefmark{2}Umeå University, Sweden \{tomas.nordstrom\}@umu.se

\IEEEauthorrefmark{3}KTH University, Sweden \{stathis,hemani\}@kth.se
}
%
%

\markboth{Journal of TBD, January~2022}%
{Shell \MakeLowercase{\textit{et al.}}: Bare Demo of IEEEtran.cls for Computer Society Journals}
%



\IEEEtitleabstractindextext{%
\begin{abstract}
 The compression of deep learning models is of fundamental importance in deploying such models to edge devices. The selection of compression parameters can be automated to meet changes in the hardware platform and application using optimization algorithms. This article introduces a
Multi-Objective Hardware-Aware Quantization (MOHAQ) method, which considers hardware efficiency and inference
error as objectives for mixed-precision quantization. The proposed method feasibly evaluates candidate
solutions in a large search space by relying on two steps. First, post-training quantization is applied for fast solution
evaluation (inference-only search). Second, we propose the "beacon-based search" to retrain
selected solutions only and use them as beacons to know the effect of retraining on other solutions. We use a speech recognition model based on Simple Recurrent
Unit (SRU) using the TIMIT dataset and apply our method to run on
SiLago and Bitfusion platforms.
We provide experimental evaluations showing that SRU can be compressed up to 8x by post-training quantization without any
significant error increase. On SiLago, we
found solutions that achieve 97\% and 86\% of the maximum possible speedup and energy saving, with a minor increase in error. On Bitfusion, beacon-based search reduced the error gain of inference-only search by up to 4.9 percentage points.
\end{abstract}

}

\maketitle

\IEEEdisplaynontitleabstractindextext

%
\IEEEpeerreviewmaketitle

\section{Introduction}
Realizing Deep Learning (DL) models on embedded platforms has become vital to bring intelligence to edge devices found in homes, cars, and wearable devices. However, DL models' appetite for memory and computational power is in direct conflict with the limited resources of embedded platforms. This conflict led to research that combines methods to shrink the models' compute and memory requirements without affecting their accuracy~\cite{rezk2019recurrent}. Models can be compressed by pruning neural network~\cite{Pruning-Dai-2017}, quantizing the weights and activations~\cite{lowp-binary-connect,HW-FPGA-FINN-L-Xilinx}, and many other methods~\cite{rezk2019recurrent}.

The effect of the model's compression can be improved by involving the target platform hardware model guidance. Different Neural Network (NN) layers have different effects on energy savings, latency improvement, and accuracy degradation~\cite{energyawarepruning-Yang-2016}. Selecting optimization parameters like pruning ratio or precision for different layers can maximize accuracy and hardware performance. Such methods are called hardware-aware optimizations. These optimization methods result in a specific solution that depends on a particular model, hardware platform, and accuracy constraint. Changes in the hardware platform or the application constraints would make this solution no more optimal. Thus, the automation of the NN compression is needed to adapt to changes in the platform and the constraints.

The compression of the NN model to run on embedded platforms can be considered an optimization problem. Some prior work on pruning and quantization has considered optimization algorithms to select pruning/quantization parameters. 
Some researchers have tried to maximize the accuracy while considering platform-related metrics such as latency or memory-size as 
constraints~\cite{postq-nahshan19,Netadapt,deepiot-Yao-2017}.
In contrast, another group of researchers considered accuracy as the constraint and selected a platform performance metric as the main objective~\cite{energyawarepruning-Yang-2016}. In this work, we treat the optimization problem as a Multi-Objective Optimization Problem (MOOP), where both the accuracy and hardware efficiency are considered objectives. First, it is non-trivial to decide whether accuracy or performance is the objective of the optimization. Thus, considering the problem as multi-objective is more appropriate. Also, MOOP has some advantages over single-objective optimization, such as giving a set of solutions that meet the objectives in different ways~\cite{MOOP-savic2002}. By doing so, the embedded system designer can do the trade-off between different alternative solutions. The role of the optimization algorithm is only to generate the most efficient solutions and not to select one solution.

However, applying multi-objective search to select NN models compression configurations is not straightforward. Many compression methods require retraining to compensate for accuracy loss, and that cannot be done during evaluating candidate solutions in a large search space. That is why in most of the recent work, the compression parameters have been selected during training and end with one solution. To make the multi-objective search possible, we rely on two aspects. First, we use post-training quantization as a compression method. Post-training quantization has become a more and more reliable compression method 
recently~\cite{postq-ocs-zhao19,postq-banner19,postq-nagel20}. Thus, the evaluation of one candidate solution would require only to run the inference of the NN model. Second, we cannot guarantee that post-training will provide high-accuracy solutions under all constraints scenarios. Thus, we propose a novel method called \textit{"beacon-based search"} to support the inference-only search with retraining. In beacon-based search, only a few solutions are selected for retraining (beacons), and they are used to guide other possible solutions without retraining all of them.

This article applies the proposed Multi-Objective Hardware-Aware Quantization (MOHAQ)  method to a recurrent neural network model used for speech recognition. Recurrent Neural Networks (RNNs) are NNs that are designed to deal with sequential data inputs or outputs applications~\cite{RNN-LSTM,rezk2019recurrent}. RNNs recognize the temporal relationship between input/output sequences by adding feedback to Feed-Forward (FF) neural networks. The model used in the experiments uses Simple Recurrent Units (SRU) instead of Long Short Term Memory (LSTM) due to its ability to be parallelized over multiple time steps and being faster in training and inference~\cite{RNN-SRU}. We selected speech recognition as an important  RNN application and the TIMIT dataset due to its popularity in speech recognition research. Also, we find good software support for speech recognition using TIMIT dataset by the Pytorch-kaldi project~\cite{pytorchkaldi-ravanelli2019}. To prove the flexibility of the method to support different applications and hardware platforms, we applied our methods to two hardware architectures with varying constraints. The first is SiLago architecture~\cite{silago-Hemani-2017}, and the second is Bitfusion architecture~\cite{bitfusion-sharma-2018}. These two architectures are chosen as they support the varying precision required in our experiments. 

The contributions of this work are summarized as:
\begin{itemize}
    \item We enable the fast evaluation of candidate solutions for multi-objective optimization for bit-width selection for NN layers by using post-training quantization. We show how to take account for combined objectives like the model error and hardware efficiency.
    \item We propose a method called beacon-based search to predict the retraining effect on candidate solutions without retraining all the evaluated solutions.
    \item We demonstrated the flexibility of the proposed method by applying it to two hardware architectures, SiLago and Bitfusion.
    \item To the best of our knowledge, this is the first work analyzing the quantization effect on SRU units. Considering SRU as an optimized version of the LSTM, it has not been investigated before if low precision quantization can be applied to SRU.
\end{itemize}

\section{Background}
\label{sec:back}
In this section, we discuss the essential components, methods, and platforms used in this work. We first explain the RNN layers related to this work. Then, we explain how NN models are quantized. Afterwards, we discuss the multi-objective optimization techniques. Finally, we cover the details of the hardware architectures used in this article.

\subsection{RNN Model Components}
RNNs recognize the temporal relation between data sequences by adding recurrent layers to the NN model. The recurrent layer adds feedback from previous time steps and memory cells. The most popular recurrent layer is called Long Short Term Memory (LSTM)~\cite{RNN-LSTM}, where the previous time-step output is concatenated with the input for the layer computations.
Here, we explain the Simple Recurrent Unit (SRU)~\cite{RNN-SRU}. SRU is an alternative for LSTM proposed to improve LSTM computational complexity. Then we briefly cover more layers required in the RNN model used in this paper experiments. 


\subsubsection{\textbf{Simple Recurrent Unit (SRU)}}
\label{subsub:qrnnsru}
Simple Recurrent Unit (SRU) is designed to make the recurrent unit easily parallelized~\cite{RNN-SRU}. Most of the LSTM computations are in the form of matrix to vector multiplications. Thus their parallelization is of great value. However, these computations rely on the previous time-step output $h_{t-1}$ and previous time-step state $C_{t-1}$ vectors, and therefore it isn't easy to be parallelized over times-steps. The SRU overcomes this problem by removing $h_{t-1}$ and $C_{t-1}$ from all matrix to vector multiplications and $C_{t-1}$ is used only in element-wise operations. The SRU is composed of two gates (forget and update gates) and a memory state. It has three matrix to vector multiplications blocks. 
    

\subsubsection{\textbf{Bidirectional RNN layer}}
In a Bidirectional RNN layer, the input is being fed into the layer from past to future and future to past. 
Obtaining data from the past and the future helps the network to understand the context better. This concept can be applied to different recurrent layer like Bi-LSTM~\cite{rnn-bi-lstm} and Bi-SRU. 

\subsubsection{\textbf{ Projection layers}}
\label{subsub:lstmproj}

The projection layer is an extra layer added before or after the recurrent layer \cite{RNN-projection}. A projection layer is similar to a Fully-Connected (FC) layer. The projection layer is added to allow an increase in the number of hidden cells while keeping the total number of parameters low. 

\subsection{Quantization}

Quantization reduces the number of bits used in neural network operations. It is possible to quantize the neural network weights only or the activations as well. The precision can change from a 32-floating point to a 16-bit fixed-point, which usually does not affect the model's accuracy. Therefore, many neural network accelerators uses 16-bit fixed point precision instead of floating point~\cite{HW-FPGA-CLSTM}. Or, quantization can be to integer precision to make it more feasible to deploy NN models on embedded platforms~\cite{post-quantize-samsung-Fang}. Integer quantization can be anything from 8 bits to 1-bit. Low precision integer accelerators have been proposed to achieve efficiency in terms of speedup, and energy and area savings~\cite{bitfusion-sharma-2018,UNPU-2019}. However, integer quantization can cause a high degradation in accuracy. Thus, retraining is required to minimize this degradation in many cases, or the model is trained using quantization-aware training from the beginning. Recently, there has been a growing interest in post-training quantization. Post-training quantization quantizes the pre-trained model parameters without any further retraining epochs after quantization. That would be useful if the training data is unavailable during the deployment time or the training platform itself is unavailable. Most post-training quantization methods work on the outlier values that consume the allowed precision and cause accuracy loss. Clipping the outliers to narrow the data range can overcome the problem, and several techniques are used for selecting clipping thresholds~\cite{postq-nahshan19,postq-ocs-zhao19}.
Alternatively, Outlier Channel Splitting (OCS) is a method that duplicates the channels with the outlier values and then halves the output values or their outgoing weights to preserve functional correctness~\cite{postq-ocs-zhao19}. However, it increases the size of the model by this channel duplication. In this work, we use clipping during quantization, and the clipping thresholds are selected using the Minimum Mean Square Error (MMSE) method~\cite{Sung_MMSE_2015}.

\subsection{Optimization using Genetic Algorithms}
Optimization is essential to various problems in engineering and economics whenever decision-making is required~\cite{optimization_chong}. It works on finding the best choice from multiple alternatives guided by an objective function and restricted by defined constraints. In many problems, there exists more than one objective. Those objectives can be conflicting that enhancing one of them requires worsening the others. Single-Objective Optimization (SOOP) will try to find a single best solution~\cite{MOOP-savic2002}. 
On the other hand, Multi-Objective Optimization (MOOP) is a kind of optimization that provides a set of solutions known as a Pareto-set (front) for conflicting objectives. Pareto-set is a non-dominating set of solutions that none of them can be further enhanced by any other solution. A solution $S1$ dominates a solution $S2$ by enhancing at least one objective without making any other objective get worse. 

Genetic Algorithms (GAs) are popular algorithms for both single and multi-objective optimization~\cite{review-GA-chahar}. GAs are inspired by natural selection, where survival is for the fittest. It is an algorithm based on populations. Each population is composed of some candidate solutions (individuals). Each individual is composed of a number of variables called chromosomes. GA is an iterative algorithm that evaluates one population solution's fitness values and generates a new population (offsprings) from the old population until a criterion is met or the defined number of populations is completed. The latest population is formed by selecting pairs of good solutions (based on their fitness values) and applying crossover on them. Crossover composes a new individual by mixing parts of parent individuals. Then mutation is applied to the offspring individual to change in some chromosomes. The selection, crossover, and mutation operations are repeated until a new population is complete. Also, an encoding function is used to map the solution variables values into another representation that can be used for genetic operations such as crossover and mutation.  

Modifications have been applied to GAs to work on multi-objective problems. These modifications are more related to fitness function assignment while the rest of the algorithm is similar to the original GA~\cite{review-GA-chahar}. There are various of multi-objective GAs such as NPGA~\cite{NPGA-Horn-1994}, NSGA~\cite{NSGA-Srinivas-2000}, and NSGA-II~\cite{NSGA2-Deb-2002}. NSGA-II (Nondominated
Sorting Genetic Algorithm II) is the enhanced version of NSGA. NSGA-II is a well-known fast multi-objective GA~\cite{NSGA2-overview-2011}. In this work, we have used the NSGA-II as our multi-objective search method.  The NSGA-II implementation is provided by PYMOO (a python library for optimization) and is based on the NSGA-II paper~\cite{NSGA2-Deb-2002}. NSGA-II is similar to a general GA, but the mating and the survival selection are modified.

\subsection{Architectures Under Study}
This section gives a brief presentation of the two architectures we have applied our methods to. The first is SiLago architecture, and the second is Bitfusion. In SiLago architecture, the low precision support is a new feature under construction. Thus, we introduce the low precision support idea in SiLago and the expected speedup and energy saving.

\subsubsection{\textbf{SiLago Architecture}} 
\label{sec:silago-back}

The architecture is a customized Coarse Grain Reconfigurable Architecture (CGRA) built upon two types of fabrics. The first fabric is called Dynamically Reconfigurable Resource Array (DRRA) and focuses on computation. The second, called Distributed Memory Architecture (DiMArch), is used as variable size streaming scratchpad memory. 

    The DRRA has a unique, extensive parallel distributed local control scheme and interconnect network \cite{sliding09}. 
    Each cell in the DRRA is comprised of a register file, sequencer, data processing unit (DPU), and switchbox. The cells in the DRRA are organized in two rows and multiple columns. The cells can communicate with each other in a 5-column span over a local network on chip (NoC)~\cite{sliding09}.  
    The DPU is using a special computation unit called Non-Linear Arithmetic Unit (NACU), operating on 16-bit words \cite{NACU20}.
    For this work, we have updated the design to be able to support three different types of low precision operations, 1x 16-bit, 2x 8-bit, or 4x 4-bit. The existing multiplier and accumulator inside the DPU were modified to use Vedic multiplication~\cite{vedic08}. 
    The DRRA is complemented by a memory fabric called DiMArch. DiMArch provides a matching parallel distributed streaming scratchpad memory \cite{DiMArch16}. 
    
    The parallel interconnect between the two fabrics makes sure that the computational parallelism is matched by the parallelism to access the scratchpad memory. 
    
    \textbf{Energy and speedup estimations} :
    Table~\ref{tab:silago-numbers} presents the energy consumption and the speedup of the arithmetic operations in the SiLago platform. The energy consumption is based on post-layout simulations of the reconfigurable multiplier and accumulator (MAC). The MAC was synthesized using a 28nm technology node. The energy consumption for the SRAM access was based on macro-generated tables in the same 28nm technology. The speedup is calculated as operations per clock cycle. The MAC can be reconfigured to calculate $1 \times$ 16-bit, $2 \times$ 8-bit, or $4 \times$ 4-bit MACs in one cycle. 
    
    \begin{table}[ht]
\centering
\caption{The speedup and energy consumed by different types of low precision operations on SiLago architecture.}

\begin{tabular}{|c||c|c|c|}
\hline


& 16x16 & 8x8 & 4x4  \\ \hline \hline

MAC speedup &1x&2x&4x \\ \hline
MAC energy cost ($pJ$) & 1.666 & 0.542  & 0.153 \\ \hline
Loading 1-bit energy cost ($pJ$) &\multicolumn{3}{c|}{ 0.08 } \\ \hline

\end{tabular}
\label{tab:silago-numbers}%
\end{table}


\subsubsection{\textbf{Bitfusion Architecture}} 
\label{sec:bitfusion-back}

Bitfusion is a variable precision architecture designed to support variations in the precision in quantized neural networks~\cite{bitfusion-sharma-2018}. It is composed of a 2-d systolic array of what is called Fused Processing Element (Fused-PE). Each Fused-PE is composed of 16 individual Bit-Bricks, each of which is designed to do 1-bit or 2-bit MAC operations. By grouping bit-bricks in one Fused-PE, higher precision operations are supported. The highest parallelism rate of one Fused-PE is 16x when the two operands are 1-bit or 2- bit, and no parallelism is achieved by having two 8-bit operands. To support 16-bit operations, the Fused-PE is used for four cycles. Thus, the speedup of using 2-bit over 16-bit operations is 64x.

%


\section{Related Work}
\label{sec:related}

In this section, we go through the related work to this article. First, we discuss the research done on the compression of SRU-based models. 
Then, we review the research relevant to the optimization of compressed neural networks.
\subsection{Simple Recurrent Unit (SRU) compression}
Shannguan \textit{\textit{et al.}} used two SRU layers in a speech recognition model as a decoder~\cite{sru-opt-shangguan19}. They managed to prune 30\% of the SRU layers without a noticeable increase in the error. The pruning was applied during training to ensure low error. 
To the best of our knowledge, no work applies quantization or any other compression method, except the pruning approaches on SRU models.

\begin{table*}[!ht]
\centering
\caption{Comparison of literature work on the optimization of NN models.}
\begin{tabular}{|c|c|c|c|c|}
\hline
 & Compression  &Objectives &Constraints & Hardware-aware\\ \hline

Yang \textit{\textit{et al.}}~\cite{energyawarepruning-Yang-2016} &Pruning&Energy&Accuracy&Energy model~\cite{energyawarepruning-Yang-2016} \\ \hline

Yang \textit{\textit{et al.}}~\cite{Netadapt}, Netadapt  &Pruning&Accuracy&Resource budget& Empirical measurements\\ \hline

Yang \textit{\textit{et al.}}~\cite{energyconstrained-Yang-2019} &Pruning&Accuracy&Energy& Energy model~\cite{energyawarepruning-Yang-2016} \\ \hline

Yao \textit{\textit{et al.}}\cite{deepiot-Yao-2017}, Deepiot &Pruning&Accuracy&Size&Memory information  \\ \hline

Rizakis\textit{\textit{et al.}} ~\cite{timeconstr-Rizakis-2018}&Pruning&Accuracy&Latency&Roofline model \\ \hline

Wang \textit{\textit{et al.}}~\cite{HAQ_Wang2019}, HAQ~&Quantization&Accuracy&Resource budget&Platform model \\ \hline

Nahshan \textit{\textit{et al.}}~\cite{postq-nahshan19}, LAPQ&Quantization&Loss&-&- \\ \hline

Cai \textit{\textit{et al.}}~\cite{postq-zeroq-cai20}, ZeroQ&Quantization&\makecell{ Size; Q.sensitivity}&Size&-\\ \hline



This work, MOHAQ&Quantization&\makecell{Error; Speedup; Energy}& Size& \makecell{Energy model~\cite{energyawarepruning-Yang-2016}; Speedup estimation} \\ \hline
\end{tabular}
\label{tab:related}%

\end{table*}

\subsection{Optimization of Neural Networks Compression}
The compression of neural network models has been treated as an optimization problem to select the degree of compression for each layer/channel. In many cases, feedback from a hardware platform or hardware model is used during the optimization (hardware-aware compression). We have summarized the work done on the optimization of NN models for compression in Table~\ref{tab:related}.
The choice of constraints and objectives have varied among different papers.

 Energy-aware pruning~\cite{energyawarepruning-Yang-2016} is a pruning method that minimizes energy consumption on a given platform. The platform model was used to guide the pruning process by informing it which layer when pruned, would lead to more energy saving. The pruning process stops when a predefined accuracy constraint has been hit.
Netadapt~\cite{Netadapt} eliminated the need for platform models by using direct empirical measurements. Nevertheless, pruning in Netadapt is constrained by a resource budget such as latency, memory size, and energy consumption. In both methods, pruning starts from pre-trained models, and fine-tuning is applied to retain accuracy. Similarly, an energy-constrained compression method~\cite{energyconstrained-Yang-2019} used pruning guided by energy constraints. Energy results are predicted from a mathematical model for a TPU-like systolic array structure architecture. However, this compression method trains the model from the beginning.
On the other hand, DeepIOT~\cite{deepiot-Yao-2017} obtains the memory size information from the target platform to compute the required compression ratio as a constraint.
In another work, optimization variables are selected based on time constraints~\cite{timeconstr-Rizakis-2018}, where roof-line models are used for calculating
the maximum achievable performance for different pruning configurations.

HAQ (Hardware-aware Quantization) used reinforcement learning to select bit-width for weights and activations to quantize a model during training while considering hardware constraints~\cite{HAQ_Wang2019}. 
 LAPQ~\cite{postq-nahshan19} and ZeroQ~\cite{postq-zeroq-cai20} applied optimization algorithms on quantized NN models but without any hardware model guidance. Loss Aware Post-training quantization (LAPQ) is a layer-wise iterative optimization algorithm to calculate the optimum quantization step for clipping~\cite{postq-nahshan19}. The authors proved that there is a relation between the quantization step and the cross-entropy loss.
Small changes in the quantization step have a drastic effect on the accuracy. LAPQ managed to quantize ImageNet models to 4-bit with a slight decrease in accuracy level.
In the Zero-shot Quantization (ZeroQ), Nagel \textit{\textit{et al.}} proposed a data-free quantization method~\cite{postq-nagel20}. Their approach uses multi-objective optimization to select the precision of different layers in the model. The two objectives are the memory size and the total quantization sensitivity, where they define an equation to measure the sensitivity of each layer for different precisions. The authors assumed that the sensitivity for each layer to a specific precision is independent of other layers' precisions. This assumption simplifies the computation of the overall sensitivity for different quantization configurations in the search space,

None of the discussed work has applied hardware-aware multi-objective optimization to the problem of NN compression. In this work, we use quantization as a compression method and target hardware models to guide the compression. We allow both the model error/accuracy and hardware efficiency metrics (speedup and energy consumption) to be objectives. We use the hardware on-chip memory size as a constraint to avoid high-cost off-chip communication. 

\section{Method}
\label{sec:method}
This section discusses the details of our proposed method for the Multi-Objective Hardware-Aware Quantization (MOHAQ) of the SRU-model for speech recognition. First, we explain how we apply the post-training quantization on the SRU. Next, we present the optimization algorithm used to select the layers' precisions guided by the hardware model. then, we explain how to enable retraining in a multi-objective search for setting mixed-precision quantization configurations. Finally, we discuss how we use the hardware models for guidance during the compression optimization.
\subsection{Post-training quantization of SRU model}
The Simple Recurrent Unit (SRU) was initially designed to overcome the parallelization difficulty in LSTM and other recurrent units. Outputs from the previous time-steps are used in the current time-step operations. This property makes it impossible to fully parallelize the M$\times$V operations over multiple time steps. In SRU, the outputs from the previous time-step are excluded from M$\times$V computations and only used in element-wise computations.

Another side effect for excluding the recurrent inputs from M$\times$V operations is that the number of weights used in the recurrent operations decreases significantly. Thus, it becomes possible to also exclude the recurrent part from low-precision quantization. We apply low-precision quantization on weights and activations used in M$\times$V operations only. Other weights are kept in a 16-bit fixed-point format. By doing so, we achieve our goal of reducing the model's overall size while keeping it performing with a low error rate. 
We have both 16-bit fixed-point and integer precisions in the same model in different layers in our work.  We here explain how quantization has been applied to the weights and activations during inference and how we move from fixed-point to integer operations and the reverse.
\begin{itemize}
\item \textbf{Weights integer quantization with clipping:}
We applied integer linear quantization on the weight matrices. We used the Minimum Mean Square Error (MMSE) method to determine the clipping threshold~\cite{Sung_MMSE_2015}. We used the implementation for CNN quantization provided by OCS paper on github~\cite{postq-ocs-zhao19} as a base for our implementations. We then modified the implementation to work with SRU units and to support varying precision per layer. The range of the quantized values are [$-128$:127], [$-8$:7], and [$-2$:1] for 8-bit, 4-bit, and 2-bit, respectively.
\item \textbf{Weights 16-bit fixed-point quantization:}
It is used for the recurrent weights, bias vectors, and weight matrices that might be chosen to have 16-bit. Depending on the range of data, we compute the minimum number of bits required for the integer part. The rest of the 16-bits are used as a sign bit and the approximated fraction part.
\item \textbf{Activation integer quantization with clipping:}
Integer quantization of activations is similar to weights. However, since we cannot compute the range of vectors required for clipping threshold computation, we calculate the expected ranges. To compute expected ranges, we first use a portion of the validation data sequences. The predicted range of a given vector is calculated as the median value of the ranges recorded while running the validation sequences. 
\item \textbf{Activation re-quantization to 16-bit fixed-point:}
The activations are quantized to 16- bits the same way as the weights. If a vector is an output of an integer operation, we found it necessary to re-quantize the values into fixed points by dividing them by a scale value. The scale value is computed to return the vector range to the same range if quantization was not applied. The range of Non-quantized data is computed using a portion of the validation data sequences while using original model weights and activation, a.k.a, turning off quantization.
\end {itemize}

\subsection{Multi-objective quantization of neural network models}
\label{sec:method-inference}
During the compression/quantization of NN models, we have two types of objectives. The first one is the NN model performance metric, such as the accuracy or the error. The second type of objective is related to the efficiency of the hardware platform, such as memory size, speedup, and energy consumption. Treating the problem as a multi-objective problem provides the designer with solutions with different options. The embedded system designer then can decide which solution is a trade-off suitable for the target application. We have used a Genetic Algorithm (GA) as it is one of the efficient multi-objective search optimizers~\cite{review-GA-chahar}. The GA we used is called NSGA-II provided by the Pymoo python library~\cite{NSGA2-Deb-2002,pymoo}. 
As mentioned in Section~\ref{sec:back}, NSGA-II is a popular GA that supports more than one objective. NSGA-II showed the ability to find better convergence and better solutions spread near the actual Pareto-optimal front for many difficult test problems~\cite{NSGA-Srinivas-2000}. Thus, NGSA-II appears to be a good candidate for our experiments.

\begin{figure}[ht]
   \centering
   \includegraphics[width=0.8\columnwidth]{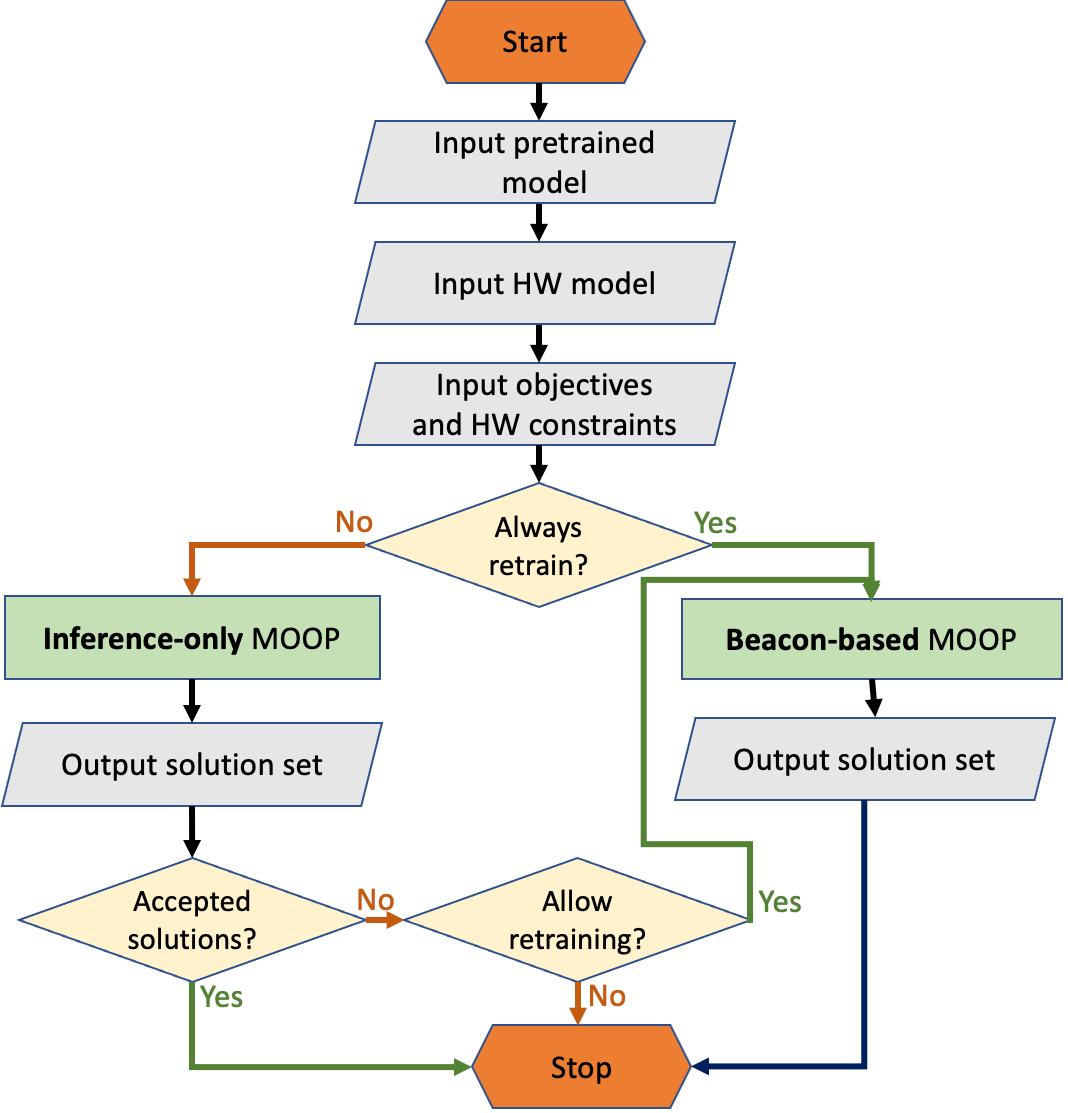}
    \caption{Steps of the Multi-Objective Hardware-Aware Quantization (MOHAQ) method. The designer inputs the model pre-trained parameters, the hardware platform objectives equations, and hardware constraints. Then, the designer can apply inference-only search or beacon-based search. The beacon-based search is the search method that requires retraining of the model using some candidate solutions variables. The output of the search is a Pareto set of optimal solutions. if the inference-only search resulted in solutions that had unaccepted accuracy levels, the designer can repeat the search using the beacon-based method.}
    \label{fig:idea}
\end{figure}

Most of the automated hardware-aware compression/quantization work use a single objective and provides a single answer. The reason is that many compression methods require training or retraining to compensate for the accuracy loss caused by compression. The selection of the compression/quantization configuration is made iteratively during training epochs. Trying to turn the problem into a multi-objective problem would make it necessary to retrain all the evaluated solutions in the search space, and that would be infeasible. Our approach to tackle this problem relies on two observations. The first is that researchers are progressively enhancing post-training compression/quantization techniques. Thus, it is possible to evaluate candidate solutions by running inference only without any retraining. That is also useful for the case when the training data is not available. The second observation is that a retrained model using one candidate solution variables can be used as a retrained model for neighbor candidate solutions in the search space. We use this observation  in a method we call \textit{"beacon-based search"} (further explained in Section~\ref{sec:beacon}). So, if the inference-only search fails to find accuracy-wise accepted solutions, the beacon-based search can be applied. alternatively, if the designer wants, the beacon-based search can be applied from the beginning and the inference-only search can be skipped. The complete search framework is demonstrated in Figure~\ref{fig:idea}.

Our problem is to select the precision for weights and activations per layer. A candidate solution has a number of variables that equals twice the number of layers. That is because each layer requires precision for the weight and precision for the activation. The possible precisions covered in this work are 2, 4, 8 bits integer, and 16-bit fixed point. We skipped precisions like 3, 5, and 6 because they are not frequently found in hardware platforms. However, other precisions can be included if required. The candidate solutions variables are encoded into genetic algorithm representation. We use discrete values for the solutions variables that are 1, 2, 3, and 4. 2- bit is encoded into 1, 4-bit is encoded into 2, 8-bit is encoded into 3, and 16-bit is encoded into 4.
In addition to encoding and decoding, we select the number of generations, define fitness functions and constraints, then keep the library default configuration for the rest of GA steps such as the crossover, mutation, and selection. For each objective, we define a fitness function. All the objectives have to be either minimization or maximization objectives. Since Pymoo by default treats objectives as minimization objectives, we change the maximization objectives to be minimization objectives by negating them.

Initially, the search relies on post-training quantization (inference-only search). The evaluation of candidate solutions does not incorporate any training, and thus it is possible to carry out the search within a reasonable time. To evaluate one solution's error objective, we run the inference of the model and get the inference error value as the error objective that we want to minimize.
We use a subset from the validation data to evaluate a quantized model. We selected half of the validation data that would resemble the same distribution of speakers' gender and dialects in the testing dataset. It should also be noted that
the evaluations of candidate solutions in one generation are not dependant on the other candidate solutions.
Therefore, it is possible to parallelize the solutions evaluation and distribute the computation over multiple GPUs~\cite{Rezk-distGA-2014} with linear speedup.

\subsection{Beacon-based Search (Retraining-based Multi-objective Quantization) }
\label{sec:beacon}

So far, we have been focusing on using post-training quantization during the search for optimum quantization configuration to achieve a speedy evaluation of candidate solutions. However, at high compression ratios, we sometimes find that the post-training quantization is not achieving acceptable accuracy levels.
The accuracy can be improved by retraining the model using the quantized parameters. Still, as retraining is a very time-consuming process, it is infeasible to do for all candidate solutions. Therefore, we have developed a beacon-based approach that only retrains a small set of solutions, our beacons. Then we let other solutions share the retrained model (beacon) as a basis for their quantization instead of the original pretrained model.

To retrain the model, we used a Binary-connect approach \cite{lowp-binary-connect}, where quantized weights are used during the forward and backward propagation only. In contrast, full
precision weights are used for the parameters update step. Therefore, in the retrained model, we always have the floating-point parameters, which can then be used as a basis for various other quantization configurations. 

We have observed that if we retrained the model using one solution low precision configuration, the retrained parameters can be used with other low precision solutions and give a similar retraining effect. And thus, we call a retrained model a beacon as it is used to predict the retraining effect on many solutions. The next question is for which solutions a beacon (retrained model) is valid and for which solutions it is not valid. We have retrained the model using multiple low precision configurations (solutions) to generate multiple beacons. For each beacon, we run a search that evaluates many solutions twice. Once using the baseline model parameters and once using the retrained model (beacon) parameters and computed, the error decrease. This experiment aims to know the borders of the beacon validity. We found that if we retrained the model using high error solution variables, the produced beacon is useful for other low precision solutions. However, in some cases, the first layer would require special precision selection in the beacon. 

The exception solutions are solutions with extremely low precision in the first layer and solutions with the first layer weights and activations in 16-bit fixed-point precision. Those two cases need special beacons crafted for them. The first layer is usually more sensitive to quantization than other layers~\cite{postq-ocs-zhao19}. Thus, this layer's very low precision needs to be fixed during retraining. However, retraining a solution with 2-bit precision in the first layer is difficult. The resulting beacon is not as good as other beacons when applied to solutions that do not have 2-bit in the first layer. Thus, we consider this a special case beacon and use it only for solutions that have 2-bit in the first layer. Also, using a beacon with the first layer parameters adapted for integer precisions is not valid if the first layer weights and activations are fixed-point. So, we create a special beacon for this case. We conclude that we need one general beacon trained on low precision solution parameters and two special beacons. One special beacon is trained using a solution that has 2-bit precision for the first layer weights. And another special beacon is trained using a solution that has 16-bit fixed-point weights and activations in the first layer.

\begin{figure*}[ht]
   \centering
    \includegraphics[width=2\columnwidth]{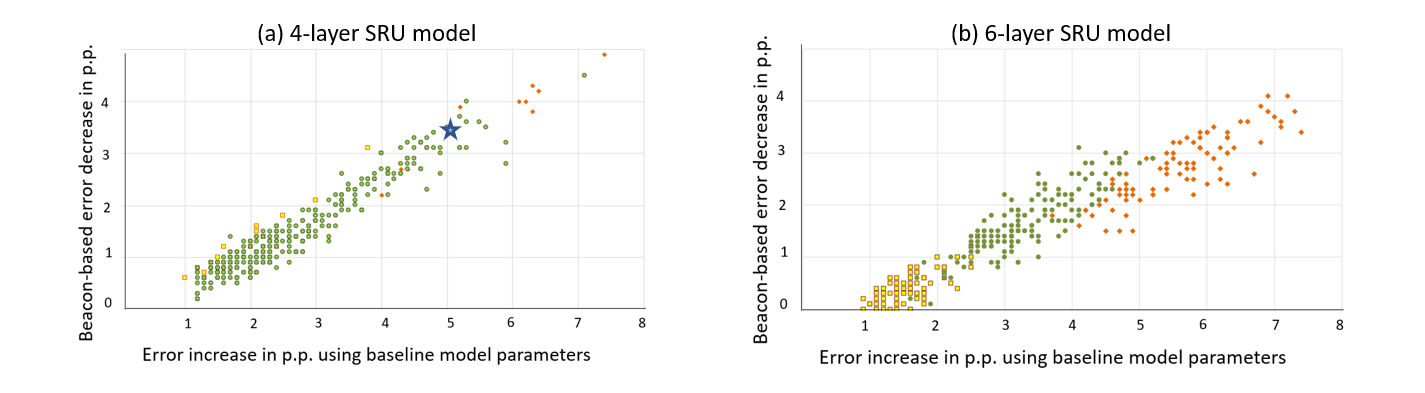}
    \caption{Plots for the error enhancement using the beacons. Each point corresponds to a solution. The x-axis corresponds to the increase in the baseline model error by evaluating solutions using baseline model parameters. The y-axis corresponds to the decrease in the error by evaluating solutions using beacons parameters. Green solutions are using the general beacon. Red solutions use the 2-bit weights first layer beacon. Yellow solutions use the fixed point first layer beacon. }
    \label{fig:beacons}
\end{figure*}

We use Figures~\ref{fig:beacons} to visualize the effect of the beacons on different solutions. In Figure ~\ref{fig:beacons}a, we use a model that has four SRU-layers, three projection layers, and one FC connected layer. In Figure~\ref{fig:beacons}b, we use a model that has six SRU layers, five projection layers, and one FC layer. In both figures, we plot the result of the experiment mentioned earlier, where we run a search that evaluates 400 solutions. The x-axis shows the increase in the error by a solution when quantizing the original model parameters compared to the Non-quantized baseline error. The y-axis shows the decrease in error when quantizing the beacon model compared to the original quantized model. If the general beacon is used, the solution is in the green color. If the solution has 2-bit weights in first layer, the solution is in red color. And, if the solution has fixed point weights and activations in the first layer, the solution is in blue color.	

To exemplify, we have marked one solution with a star in Figure~\ref{fig:beacons}a. For this solution, the baseline model error rate is 17.2\%, and applying post-training quantization on this solution using the baseline model parameters gives an error rate of 22.2\%. Applying post-training quantization on the same solution using the beacon parameters gives an error rate of 18.7\%.	
Thus, we find the particular solution at 5 ($22.2-17.2$) on the x-axis and 3.5 ($22.2-18.7$) on the y-axis.	
The figure shows close to a linear relationship between the increase in the error using the baseline model parameters and the decrease in the error using the beacons parameters. We conclude that there is no need to retrain the model for all evaluated solutions during the search. It is sufficient to train a small set of solutions and use them as beacons to predict the effect of retraining on other solutions. If the solution is feasible during solutions evaluation, we evaluate it twice, once using the baseline model parameters and once using the beacon parameters. The minimum of the two errors is used as the error objective value.

We use the term beacon due to the similarity to its use in swarm robots search, where the simple robots do not have the communication capabilities needed for the swarm to fulfill its task and some of the robots are assigned to be fixed communication beacons for the others~\cite{swarm-beacons-2013}. Similarly, when our search reaches an area where applying quantization on baseline model parameters leads to high error, a beacon is used to retrieve some of the accuracy loss. By the end, we get a solution-set that considers the retraining effect on the candidate solutions. Later on, when the designer selects a solution from the Pareto optimal set, the designer can use the beacon parameters directly or retrain the model using the selected solution parameters.

\subsection{Hardware-aware Optimization}

\label{sec:HW-aware}
In this work, the hardware model is an input given to the optimization algorithm in the form of objective functions. We have selected two architectures to apply our methods on. These two architectures are selected as they support varying precision operations, and thus applying quantization optimization becomes feasible. For these two architectures we do not have implementations for the RNN modules and thus we cannot get empirical measurements during evaluations. Instead, we have defined objective functions in a simple way that mainly focuses on the effect of decreasing the precision of the NN operations. The hardware platform constraints and objectives serve as a proof of concept that shows how models can be compressed differently.

Neural network models are characterized by having large memory requirements. Deploying the models in their original form results in frequent data loading from and into the off-chip memory. Thus, such implementation would be memory bounded, and many studies have been performed to increase the reuse of local data and minimize the off-chip memory data usage. On the other hand, the success of NN compression has made it possible to squeeze the whole NN model into the on-chip memory and transfer the NN applications into compute-bound applications. So, in our experiments, we use the platform SRAM size as an optimization constraint and not an objective. First, having the NN model size less than the SRAM size achieves the ultimate goal of compression by avoiding the expensive loading of weights from the off-chip memory. Second, compressing the model more would not be beneficial anymore from the memory point of view. But, it still can be beneficial for energy consumption and computation speedup, which are accounted for as optimization objectives. Next, we show the details of the energy and speedup objectives equations.

\subsubsection{Energy Estimation}
In our experiments, we use the energy estimation model developed by the Eyeriss project team~\cite{energyawarepruning-Yang-2016}. In this model, the total consumed energy is computed by adding the total energy required for computation and the total energy required for data movements. Since the majority of computation in NN models are in the form of MAC operations, the total energy required for computation can be approximated as the number of MAC operations multiplied by the energy cost of one MAC operation. The total energy consumed by data movement is computed by multiplying the cost of one-bit transfer by the number of transferred bits.



\subsubsection{Speedup Estimation}
 
Since we adopt compute-bound implementations in this work, we rely solely on the speedup gained at the MAC operations computations as an approximation for the expected speedup gained by different quantization configurations.
In both architectures, the highest supported precision is the 16-bit fixed point. Thus, we define an objective for the speedup to compute the speedup over 16-bit operations. The speedup objective is computed by multiplying the number of MAC operations done in a given precision by the speedup of this precision and sum over all supported precisions using the following formula:

\begin{equation}
\label{eq:speedup}
 S = \sum_{i=0}^{|P|} \frac{S_i*N_i}{N_T} ~~~,
\end{equation}

where $S$ is the overall speedup, and $P$ is the set of supported precisions. 
For example, an architecture that supports mixed-precision with 4 and 8 bits have a set $P= \{8*8, 4*8, 4*4\}$, with  $|P|= 3$. If the same architecture does not support mixed-precision,  $P=\{8*8, 4*4\}$, with $|P|= 2$. 
Furthermore, $S_i$ is the speedup gained using precision $p_i$ over the highest precision supported by the given architecture. $N_i$ is the number of MAC (Multiply-Accumulate) operations using the precision $p_i$, and $N_T = \sum_{i}N_i$ is the total number of MAC operations in the model.

\section{Evaluation and Experiments}
\label{sec:evaluate}
This section applies the Multi-Objective Hardware-Aware Quantization (MOHAQ) method using the NSGA-II genetic algorithm to an SRU model using the TIMIT dataset for speech recognition. As we mentioned in the introduction, SRU-model is selected due to the ease of SRU parallelization and for being faster in training and inference than other recurrent layers. Also, we found in the Pytorch-Kaldi project good software support for the SRU-based models for speech recognition using TIMIT dataset~\cite{pytorchkaldi-ravanelli2019}. TIMIT is a dataset composed of recordings for 630 different speakers using six different American English dialects, where each speaker is reading up to 10 sentences~\cite{dataset-timit}. In Section~\ref{sec:model} we show the components of the model used in our experiments.

We designed three experiments to evaluate our MOHAQ method. In the three experiments, the search should give a set of Pareto optimum solutions. Each solution is the precision of each layer and activation in the model. In the first experiment, we evaluate the capabilities of the post-training quantization on the SRU model without any hardware consideration. In the latter two experiments, we use two hardware models, SiLago and Bitfusion. SiLago does not support precision less than 4-bit; inference-only search was enough for the example model as the compression ratio did not exceed 8x.
On the other hand, Bitfusion supports 2-bit operations and hence supports high compression ratio solutions. Therefore, this example architecture gave us an opportunity to test the search method at a high compression ratio by setting the memory constraint to 2 MB (10.6x compression ratio). We first apply the inference-only search, and then we use the beacon-based search and examine the quality of the solution set. Afterwards, we repeated the experiment for a deeper model that has 4 more layers to test the scalability of the method. In this case, we set the memory constraint to be 3MB to suit the deeper model.

\subsection{Example SRU Model}
\label{sec:model}
In our experiments, we use a speech recognition model from the Pytorch-Kaldi project~\cite{pytorchkaldi-ravanelli2019}. Pytorch-Kaldi is a project that develops hybrid speech recognition systems using state-of-the-art DNN/RNN. Pytorch is used for the NN acoustic model. Kaldi toolkit is used for feature extraction, label computation, and decoding~\cite{kaldi_ASRU2011}. In our experiments, the feature extraction is done using logarithmic Mel-filter bank coefficients (FBANK). The labels required for the acoustic model training come from a procedure of forced alignment between the context-dependent phone state sequence and the speech features. Then, the Pytorch NN module takes the features vector as input and generates an output vector. The output vector is a set of posterior probabilities over the phone states. The Kaldi decoder uses this vector to compute the final Word-Error-Rate (WER).
We used the TIMIT dataset~\cite{dataset-timit} and trained the model for 24 epochs as set in the Pytorch-Kaldi default configurations.

\begin{figure}[ht]
   \centering
   \includegraphics[width=0.90\columnwidth]{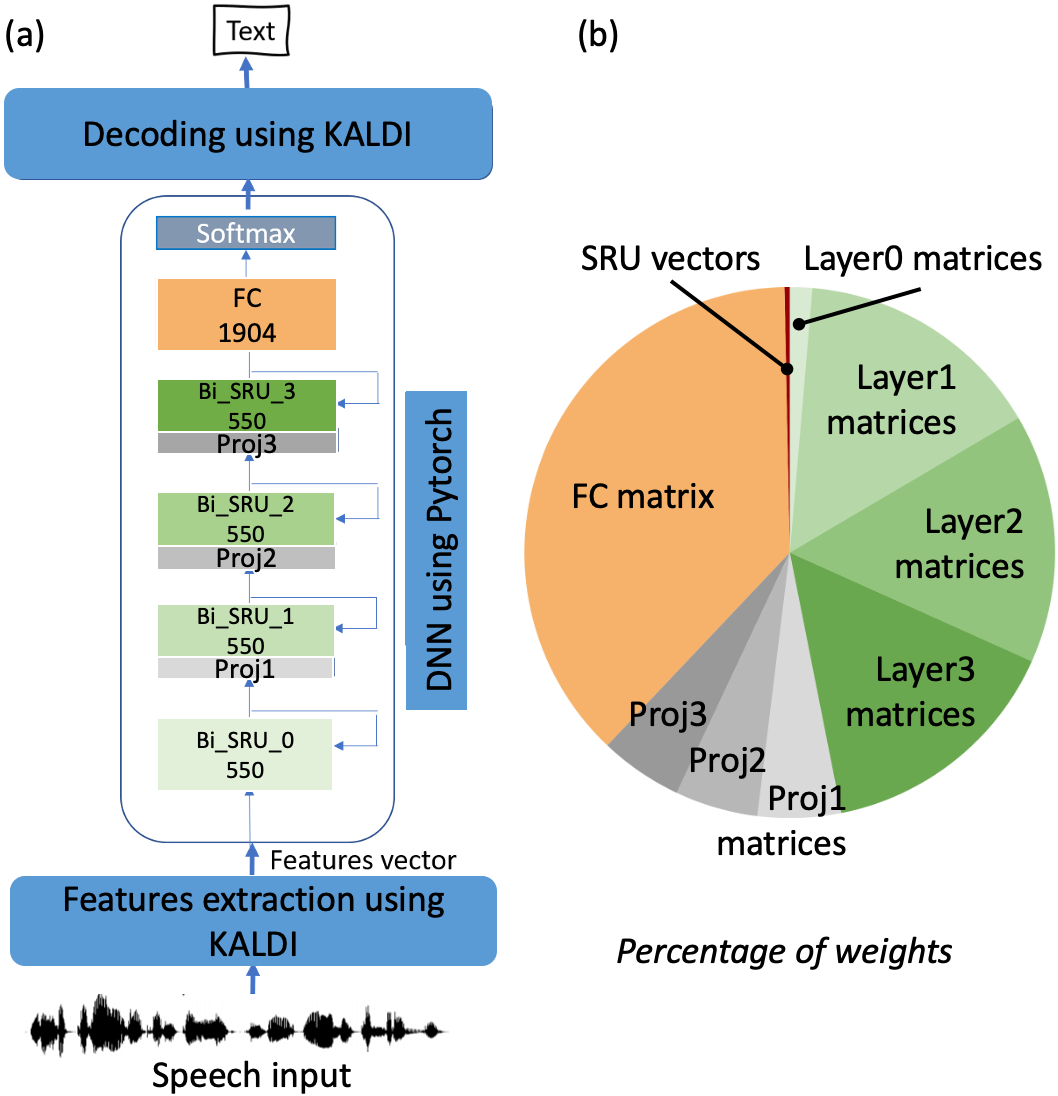}
    \caption{Model understudy: 
    (a) The ASR model using KALDI for features extraction and decoding and SRU-based RNN for speech recognition;
(b) A pie chart showing the percentage of weights allocated in memory to each layer in the SRU-model understudy. SRU matrices layers are in green. Projection matrices are in grey. FC matrix is in orange and SRU vectors are in red.
    }
    \label{fig:model}
\end{figure}

Figure ~\ref{fig:model}a shows the model used in our experiments. The NN model is composed of 4 Bi-SRU layers with 3 projection layers in between. A FC layer is used after the SRU layers and the output is applied to a Softmax layer~\cite{pytorchkaldi-ravanelli2019}. In Figure~\ref{fig:model}b, we show the percentage of weights required by each layer. 

.
\subsection{Multi-objective search to minimize two objectives: WER and memory size}
\label{sec:per-exp}
\label{sec:wer-mem}
\begin{figure}[ht]
   \centering
   \includegraphics[width=1\columnwidth]{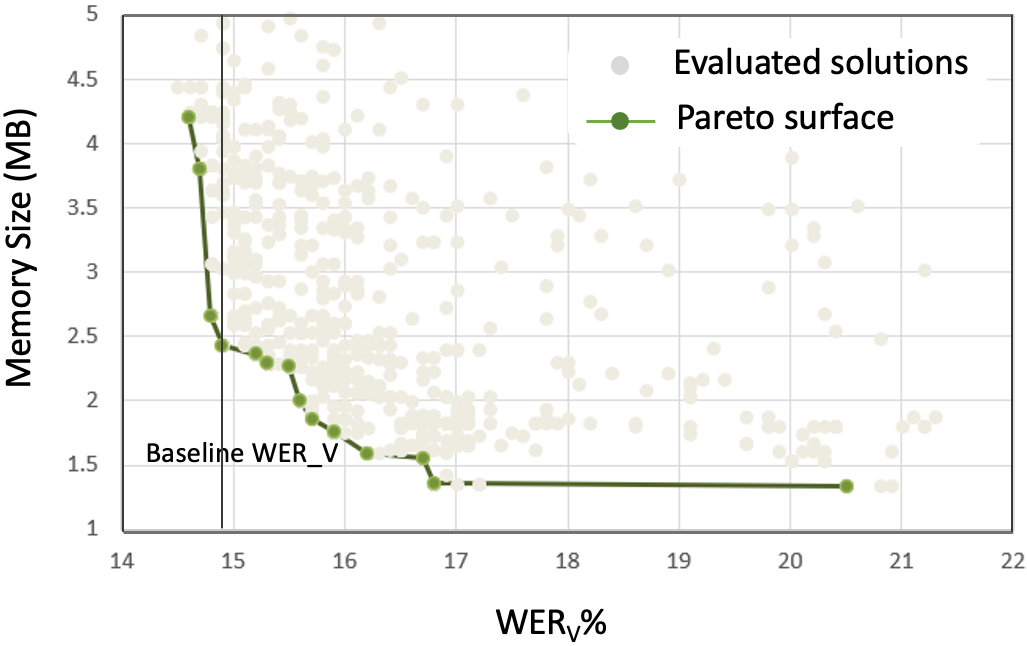}
    \caption{Pareto optimal set for two objectives: validation error ($WER_V$) and memory size. }
    \label{fig:mem-wer}
\end{figure}

Our first experiment does a multi-objective search to minimize two objectives: $WER_V$ and memory size. $WER_V$ is the error rate evaluated using the validation set of the TIMIT dataset. No hardware model is used in this experiment to explore the general compression of the model before any hardware platform is involved. In the search space, we have 4.3 billion possible solutions ($4^{16}$) as each solution has 16 variables, and each variable has four possible values. The genetic algorithm used is NSGA-II. After some initial experiments, we found that 60 generation was sufficient to get a stable solution for all tested objectives. Each generation has ten individuals, except the initial generation, which has 40 individuals. Thus, 630 solutions have been evaluated during the search. Note that, high error rate solutions are infeasible. The search output is a Pareto optimal set of solutions that shows a trade-off between the model size and the error rate to the embedded system designer. Figure~\ref{fig:mem-wer} shows a plot for the Pareto optimal set, and Table~\ref{tab:wer-mem} shows the details of each solution in the set. We report the precision of each layer weight and activation for each solution, followed by the solution $WER_V$, compression ratio, and testing error $WER_T$. The first row is for the base model that is not quantized. The base model $WER_T$ is $17.2\%$. 

Table~\ref{tab:wer-mem} shows that the model can be compressed to 8.7x without any increase in the error. The designer can compress the model to 12x with only 1.2 p.p. (percentage point) increase in the error and to 15.6x with 2.1 p.p. increase in error. Increasing the compression ratio to its maximum value (15.8x) leads to a significant increase in the error, as  seen when solutions $S17$ and $S18$ are compared. The difference between the errors in these solutions is high, but the difference between the memory sizes is slight. Thus, there is no need to further compress the model as in $S18$.

In most of the solutions, 4 bits and 2 bits have been used extensively for the weights. The activation precision has been kept between 8-bit and 16-bit in some layers for most of the solutions. However, 4-bit and 2-bit activations have been used in few layers.
 It is also observed that some solutions have an error rate better than the baseline model. It has been shown that quantization has a regularization effect during training~\cite{HW-FPGA-FINN-L-Xilinx}. Therefore, we think the improved error is a result of the quantization error introducing a noise that reduces some of the over-fitting effect during inference. 

In Table~\ref{tab:wer-mem}, we expected the $WER_T$ to be higher than $WER_V$ but also we hoped to see that the relative order is kept between the solutions for both $WER_T$ and $WER_V$. However, if we look at the solutions sorted by the $WER_V$ value, we find that the corresponding $WER_T$ values are not perfectly sorted. $S6$ and $S12$ look as outliers. The reason is that $S6$ and $S12$ had better $WER_T$ than expected. Still, for both cases, the variation was in the range of 0.2 p.p. and we believe that such small variations are expected as there is no guarantee for two different datasets' errors to be the same. 


\begin{table*}[!ht]
\centering
\caption{ The Pareto-set of solutions resulted from applying NSGA-II minimizing two objectives: $WER_V$ and memory size in MB. $WER_V$ is the error rate evaluated using the validation set. The  testing error is previewed in the table as $WER_T$. Each layer is denoted as $L_x$, and each projection layer is denoted as $Pr_x$, where x is the layer index. For each layer, the number of bits used is written as W/A, where W is the number of bits for weights and A is the number of bits for activations. $Cp_r$ is the compression ratio. The base model is a full precision implementation with $WER_T$ of 17.2\%.}
\begin{tabular}{|c|c|c|c|c|c|c|c|c|c|c|c|}
\hline
Sol.&$L_0$ & $Pr_1$&$L_1$& $Pr_2$& $L_2$& $Pr_3$ &$L_3$& FC & $WER_V$ & $Cp_r$&
$WER_T$\\ \hline\hline

Baseline&32/32&32/32&32/32&32/32&32/32&32/32&32/32&32/32&14.9\% & 1x&17.2\% \\ \hline\hline

S1&8/16&	16/4	&4/4	&4/16&	4/16	&8/4	&16/16	&4/8&	14.5\%&	4.8x&	17.1\% \\ \hline

S2&4/16	&2/16	&4/4	&4/4&	4/16	&16/8	&16/16	&4/8	&14.6\%&	5.0x&	17.1\% \\ \hline

S3&4/8&	2/4&	4/4	&4/8&	4/8	&16/8	&16/16&	4/8&	14.6\%&	5.0x&	17.2\% \\ \hline

S4&4/8	&2/16	&4/4	&4/8&	4/16	&4/8	&16/16	&4/8&	14.7\%&	5.6x&	17.2\% \\ \hline

S5&4/16	&2/4	&4/4	&4/8&	4/16	&4/16	&16/16	&4/8	&14.7\%&	5.6x&	17.2\% \\ \hline

S6&4/8	&2/4&	4/4	&2/4	&4/16&	8/16	&4/16&	4/8&	14.8\%&	8.0x& 17.0\% \\ \hline

S7&8/16	&4/16	&4/4	&4/16&	2/16&	2/8&	4/16	&4/8	&14.9\%&	8.7x&	17.2\% \\ \hline

S8&8/16	&4/4&	4/4	&2/16&	2/16	&2/16&	4/16&	4/8&	15.2\%&	9.0x&	17.3\% \\ \hline

S9&8/8	&2/4	&4/4	&2/16&	2/16	&2/16	&4/16	&4/4&	15.3\%&	9.2x&	17.3\% \\ \hline

S10&8/16	&2/4	&4/4	&2/8&	2/16	&2/8	&4/16	&4/4&	15.3\%&	9.2x&	17.4\% \\ \hline

S11&8/16	&4/4	&4/4	&2/16&	2/16	&2/8	&8/16	&2/4	&15.5\%&	9.4x&	18.2\% \\ \hline

S12&16/16	&4/16	&4/4	&2/16&	2/16	&2/8	&4/16	&2/8	&15.6\%&	10.6x&	18.0\% \\ \hline

S13&8/8&	4/4	&4/4	&2/8&	2/16	&2/8&	4/16	&2/4&	15.7\%&	11.4x&	18.4\% \\ \hline

S14&4/8&	2/2&	4/4	&2/16&	2/16	&2/16	&4/16	&2/8	&15.9\%&	12.0x&	18.4\% \\ \hline

S15&8/16&	2/2	&2/4	&2/8	&2/16	&2/8&	4/16	&2/4&	16.2\%&	13.2x&	19.1\% \\ \hline
								
S16&4/8&	2/2	&4/4	&2/4	&2/16	&2/8&	2/16	&2/8&	16.7\%&	13.6x&	19.2\% \\ \hline
								
S17&4/16&	2/2	&2/4	&2/8	&2/16	&2/8&	2/16	&2/8&	16.8\%&	15.6x&	19.3\% \\ \hline
								
S18&2/8&	2/2	&2/4	&2/4	&2/16	&2/8&	2/16	&2/8&	20.5\%&	15.8x&	23.6\% \\ \hline

\end{tabular}
\label{tab:wer-mem}%

\end{table*}

To get a better understanding of how successful our post-training quantization applied to the SRU-model is, we look at the previous work done on post-training quantization. Since researchers have found that 16-bit and 8-bit quantization do not significantly affect accuracy~\cite{postq-ocs-zhao19}, we will focus on 4-bit quantization (8x compression). CNN ImageNet models have been used for quantization experiments in most of the work we have seen. The accuracy drop due to 4-bit post-training quantization on CNN models has varied in these papers as follows: LAPQ~\cite{postq-nahshan19} (6.1 to 9.4 p.p.), ACIQ~\cite{postq-banner19} (0.4 to 10.8 p.p.), OCS~\cite{postq-ocs-zhao19} (more than 5 p.p.), and ZeroQ~\cite{postq-zeroq-cai20} (1.6 p.p.). Where ZeroQ applied mixed precision to reach the 8x compression ratio. Also, on RNN models for language translation, the BLEU score decreased by 1.2~\cite{NMTquant-Aji2019}. Comparing this to the mixed-precision post-training of the SRU model, our search found solutions that use a mix of 2, 4, and 8-bits. Those solutions achieve compression ratios between 8x and 9x with a negligible error increase. With an error increase of 1.2 p.p., the compression ratio reaches 12x. Thus, we conclude that the error increase we get is lower than most of the other studies, and we can see that we have higher compression ratios in our experiments.

\subsection{Multi-objective Quantization on the SiLago architecture }
\label{sec:silago-exp}
\label{sec:silago-ev}
In the second experiment, we apply the MOHAQ method for SiLago Architecture using the inference-only search. This architecture can support varying precisions between layers. However, for each layer, the weight and the activation must use the same precision. Thus, the number of variables in a solution is 8, not 16, as in the previous experiment. The precisions supported on SiLago are 16, 8, and 4 bits, as explained in Section~\ref{sec:silago-back}. Since the highest precision supported on SiLago is a 16-bit fixed point, the baseline model is a 16-bit fixed-point implementation,
towards which the speedup is compared using Equation~\ref{eq:speedup}.
%
Also, we have evaluated the energy consumption required by the MAC operation for different precisions. Table~\ref{tab:silago-numbers} shows the speedup gained per  MAC operation when using 8 or 4-bit when compared to 16-bit and the energy consumed by 16, 8, and 4-bit operations. To compute the expected overall required energy by each solution, we use the energy model proposed by Energy-aware pruning~\cite{energyawarepruning-Yang-2016} and described in Section~\ref{sec:HW-aware}.

As discussed in Section~\ref{sec:HW-aware}, all weights should be stored in the on-chip memory. Thus, it is crucial to add the SRAM size as a constraint for the model size during the search. Even though we do not know the exact size of the SRAM, we have to establish a limit. Given that the highest possible compression ratio on SiLago is 8x, which corresponds to 2.65 MB for the experimental model, we chose 6 MB (3.5x compression ratio) as a reasonable memory size constraint.
The multi-objective search has three objectives: WER, speedup, and energy consumption. Both WER and energy consumption are objectives to be minimized, while speedup is an objective to be maximized. We use the negative of the speedup as an objective instead of the speedup as the GA will minimize all the objectives.

Since the search space is smaller than the search space in the first experiment, we only need to run the search for 15 generations. Each generation has ten individual solutions, except the first generation has 40 individual solutions. The whole search, therefore, evaluates 180 solutions out of 6561 possible solutions ($3^8$). Solutions with a high error rate were considered infeasible. 
Table~\ref{tab:silago} shows the Pareto optimal set with each solution details and the testing WER. To judge the speedup and energy consumption quality, we compare the solutions against the best possible performing solution on SiLago, which is using 4-bit for all layers. This solution can reach a 3.9x speedup and the lowest energy consumption of $2.6\mu J$ (a 6.3x improvement than the base solution).
The search managed to find solutions that can achieve 94\% of the maximum speedup and 70\% of the maximum energy saving without any increase in the error. If the designer agreed to have a 0.3 p.p. increase in the error, we could achieve 97\% of the maximum speedup and 86\% of the maximum energy saving. 

\begin{table*}[!ht]
\centering
\caption{ The Pareto-set of solutions resulted from applying NSGA-II minimizing three objectives, $WER_V$ and speedup and energy consumption. $WER_V$ is the error rate evaluated using the validation set. The speedup and energy consumption are computed using SiLago architecture model. The testing error is also previewed in the table as $WER_T$. Each layer is denoted as $L_x$, and each projection layer is denoted as $Pr_x$, where $x$ is the layer index. For each layer, the number of bits used is written as $W/A$, where $W$ is the number of bits for weights and $A$ is the number of bits for activations. $Cp_r$ is the compression ratio. $Base_S$ is the base model that can run on SiLago using a 16-bit full implementation.  }
\begin{tabular}{|c|c|c|c|c|c|c|c|c|c|c|c|c|c|c|c|}
\hline
Sol.&$L_0$ & $Pr_1$&$L_1$& $Pr_2$& $L_2$& $Pr_3$ &$L_3$& FC & $WER_V$ & $Cp_r$& Speedup&  Energy &
$WER_T$\\ \hline\hline
Base&32/32&32/32&32/32&32/32&32/32&32/32&32/32&32/32&14.9\% & 1x&-&-&17.2\% \\ \hline\hline
$Base_S$&16/16 & 16/16 &16/16&16/16&16/16&16/16&16/16&16/16&14.9\% & 2x& 1x&16.4 \textmu J&17.2\% \\ \hline\hline

S1&8/8&	4/4	&4/4&16/16	&4/4	&8/8	&16/16&	8/8	&14.4\% & 3.9x&	2.5x&	7.2 \textmu J&16.9\% \\ \hline

S2&16/16&	8/8	&4/4	&4/4	&4/4	&16/16&	8/8	&4/4&14.5\% & 5.7x&	3.3x&	4.3 \textmu J&	17\% \\ \hline

S3&16/16	&4/4	&4/4	&4/4&	4/4	&16/16	&8/8  &4/4	&14.6\% &5.7x&	3.4x&	4.1 \textmu J&	16.8\% \\ \hline

S4&16/16	&8/8	&4/4	&8/8 &4/4	&16/16		&4/4&	4/4	&14.9\% & 5.8x&	3.5x&	3.9 \textmu J&17.2\% \\ \hline

S5&16/16	&4/4	&4/4	&8/8 &4/4	&16/16		&4/4&	4/4&	15.1\% & 6.6x&	3.6x&	3.7 \textmu J&	17.2\% \\ \hline

S6&16/16	&4/4	&4/4	&8/8	&4/4	&4/4	&4/4&	4/4	&15.3\% &7.3x&	3.8x&	3 \textmu J&	17.5\% \\ \hline

\end{tabular}
\label{tab:silago}%

\end{table*}


\subsection{Multi-objective Quantization on the Bitfusion architecture }
\label{sec:bitfusion-exp}
In the third experiment, we apply the Multi-Objective Hardware-Aware  Quantization (MOHAQ) method for the Bitfusion architecture using two objectives. The first is the WER and the second is the speedup. We first apply the inference-only search, and then we apply the beacon-based search to enhance the solution set. Bitfusion architecture is introduced in Section~\ref{sec:bitfusion-back}. We use Equation~\ref{eq:speedup} to compute the expected speedup for different solutions. The genetic algorithm used is NSGA-II, and it runs for 60 generations as done in the first experiment. Each generation has ten individuals except the initial generation, which has 40 individuals. The whole search evaluates 630 solutions out of 4.3 billion possible solutions ($4^{16}$). To show the scalability of the beacon-based search method, we repeated the experiment for a deeper model that has 6 SRU layers, 5 projection layers, and 1 F.C. layer. The new model is constructed by duplicating SRU and projection layers in the original model. The duplicated layers are similar to layers 1, 2 and 3. The new model has 12 layers and 24 search variables (2 variables per layer). Therefore, we run the search for 90 generations for the second model to suit the enlarged search space.

We considered high error solutions (more than 24\%) infeasible to limit the search to low error rate solutions only.
The speedup equation assumes that all the weights have to be in the SRAM, and the application is compute-bound. Thus, memory size has to be a constraint in the search.
During the design of this experiment, we put a constraint for the memory size to be very small to allow the search to find high error solutions to do the beacon-based search as a next step. In the first model, the memory constraint is 2 MB, and for the second model, it is 3 MB. These memory sizes are equivalent to 9.4\% and 10.1\% of the original models' sizes, respectively.

\begin{figure*}[ht]
   \centering
   \includegraphics[width=1.95\columnwidth]{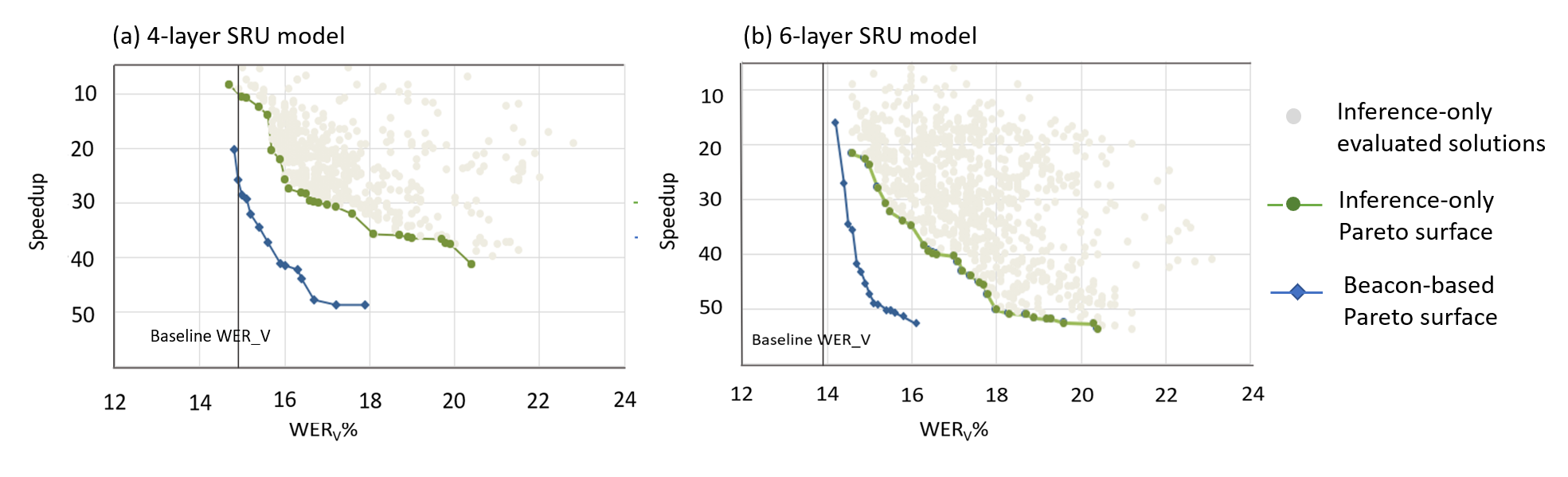}
    \caption{Pareto optimal set for two objectives, validation error ($WER_V$) and speedup on Bitfusion architecture model. We compare the two Pareto sets generated by the inference-only search (in green) and the beacon-based search (in blue). Evaluated solutions during the inference-only search are also previewed (in gray).}
    \label{fig:bitfusion}
\end{figure*}

In Figure~\ref{fig:bitfusion}, we show the experiment results for the 4-layer SRU model and the 6-layer SRU model, respectively. The figures show the inference-only search Pareto-front in green and the evaluated solutions in gray. After repeating the search using the beacon-based search, the Pareto-front is plotted in blue. The gap between the two Pareto sets in both figures is the outcome of applying retraining via beacons. 
We present the details of the Pareto sets for the 4-layer SRU model in Table~\ref{tab:bitfusion} and for the 6-layer SRU model in Table~\ref{tab:bitfusion}. We excluded the solution variables values in these tables, and skipped some solutions to keep the tables concise.

Comparing the two Pareto sets in Table~\ref{tab:bitfusion}, the testing error on the first set reached 23.4\% to achieve a 41x speedup in $SI25$. The new set reached the same speedup level with a testing error of 18.5\% in $SB10$. The new set also found more solutions with higher speedups up to 48.8x with a testing error of 20.3\%.
If we compare the two Pareto sets for the deeper model in Table~\ref{tab:bitfusion-6}, we find that the testing error on the first set reached 20.9\% to achieve a 50x speedup in $SI20$. The new set reached the same speedup level with a testing error of 18\% in $SB12$. Also, the testing error on the first set reached 22.3\% to achieve a 52.6x speedup in $SI26$. The new set reached the same speedup level with a testing error of 18.6\% in $SB16$. 

We have noticed that at high compression ratios, the relative order of testing error of solutions sorted by their validation error is not as good as in less-compressed solutions as in Sections~\ref{sec:wer-mem} and~\ref{sec:silago-ev}. For example, in Table~\ref{tab:bitfusion-6}, solutions $SI12$ and $SI16$ have the same testing error despite having a difference of 0.8 p.p. between their validation errors.  



\begin{table}[!ht]
\centering
\caption{The Pareto-sets of solutions resulted from applying the NSGA-II algorithm minimizing two objectives, $WER_V$ and speedup on the 4-SRU model. $WER_V$ is the error rate evaluated using the validation set. The speedup is computed using Bitfusion architecture model. The  testing error is previewed in the table as $WER_T$.  $Cp_r$ is the compression ratio. $Base_F$ is the base model that can run on Bitfusion using a 16-bit  implementation. Solutions start with $SI$ belongs to the inference-only search Pareto set and solutions start with $SB$ belongs to the beacon-based search.}
\begin{tabular}{|c|c|c|c|c|}
\hline
Sol. & $WER_V$ & $Cp_r$&Speedup&
$WER_T$\\ \hline\hline

Base&14.9\% & 1x&-&17.2\% \\ \hline\hline
$Base_F$&14.9\% & 2x& 1x&17.2\% \\ \hline\hline

S1&	14.7\%&	5.6x&	8.3x&	17\% \\ \hline


SI5&	15.6\%&	7.5x&	13.8x&	18.3\% \\ \hline
SI8&	16\%&	7.4x&	25.7x&	18.6\% \\ \hline
SI14&	16.8\%&	11.4x&	29.9x&	19.2\% \\ \hline
SI17&	17.6\%&	11.4x&	32.1x&	20.3\% \\ \hline
SI23&	19.8\%&	13.8x&	37.5x&	21.3\% \\ \hline
SI25&	20.4\%&	13.3x&	41.3x&	23.4\% \\ \hline \hline

SB1&	14.8\%&	11.8x&	20.2x&	17.5\% \\ \hline
SB2&	14.9\%&	11.8x&	25.7x&	17.7\% \\ \hline
SB5&	15.2\%&	11.8x&	32.1x&	18.2\% \\ \hline
SB7&	15.6\%&	13.3x&	37.3x&	18.8\% \\ \hline
SB10&16.3\%&	14.5x&	42.3x&	18.5\% \\ \hline
SB12&16.7\%&	14.5x&	47.9x&	19.5\% \\ \hline
SB14&17.9\%&	15.6x&	48.8x&	20.3\% \\ \hline

\end{tabular}
\label{tab:bitfusion}%

\end{table}

\begin{table}[!ht]
\centering
\caption{The Pareto-sets of solutions resulted from applying the NSGA-II algorithm minimizing two objectives, $WER_V$ and speedup on the 6-SRU model. $WER_V$ is the error rate evaluated using the validation set. The speedup is computed using Bitfusion architecture model. The  testing error is previewed in the table as $WER_T$. $Cp_r$ is the compression ratio. $Base_F$ is the base model that can run on Bitfusion using a 16-bit  implementation. Solutions start with $SI$ belongs to the inference-only search Pareto set and solutions start with $SB$ belongs to the beacon-based search.}
\begin{tabular}{|c|c|c|c|c|}
\hline
Sol. & $WER_V$ & $Cp_r$&Speedup&
$WER_T$\\ \hline\hline

Base&13.9\% & 1x&-&16.1\% \\ \hline\hline
$Base_F$&13.9\% & 2x& 1x&16.1\% \\ \hline\hline

SI1&	14.6\%&	10.3x&	21.6x&	17.3\% \\ \hline
SI4&	15.2\%&	11.6x&	28x&	17.5\% \\ \hline
SI8&	16\%&	11.6x&	34.9x&	18.0\% \\ \hline
SI12&16.6\%&	11.8x&	40.1x&	19.5\% \\ \hline
SI16&	17.4\%&	15.2x&	44.0x&	19.5\% \\ \hline
SI20&	18.0\%&	15.7x&	50.2x&	20.9\% \\ \hline
SI23&	18.9\%&	15.7x&	51.7x&	21.0\% \\ \hline
SI26&	19.6\%&	14.9x&	52.6x&	22.3\% \\ \hline\hline

SB1&	14.2\%&	10.3x&	16.0x&	16.9\% \\ \hline
SB2&	14.4\%&	10.0x&	27.1x&	17.0\% \\ \hline
SB3&	14.5\%&	10.5x&	34.6x&	17.2\% \\ \hline
SB6&	14.8\%&	12.8x&	43.3x&	17.5\% \\ \hline
SB9&	15.1\%&	13.1x&	48.9x&	18.2\% \\ \hline
SB12&15.5\%&	12.0x&	50.3x&	18.0\% \\ \hline
SB16&16.1\%&	13.5x&	52.6x&	18.6\% \\ \hline
\end{tabular}
\label{tab:bitfusion-6}%

\end{table}

\section{Discussion and Limitations}

The main focus of this paper is to open up a research direction for \textit{hardware-aware multi-objective compression} of neural networks. There exist many network models and a large number of compression techniques. In this work, we principally focus on quantization and application to SRU models for several reasons. Quantization is one of the vital compression methods that can be used solely or with other compression methods. Therefore, we consider enabling the MOOP on quantization is of great benefit. Also, SRU is a promising recurrent layer that allows a powerful hardware parallelization. An additional reason for using SRU is to examine to what extent it can be quantized as it is not well understood. 
Since the SRU can be considered as an optimized version of the LSTM, we wanted to investigate the effect of quantization on it. Our experiments showed that by excluding the recurrent vectors and biases from quantization, SRU could be quantized to high compression ratios without a harsh effect on the model error rate. Thus, the SRU combines both the high parallelization speedup and compression model size reduction benefits. The second reason is that running experiments using SRU is much faster than other RNN models. Thus, we had a better opportunity to have multiple trials to explore our methodology.

In this article, we further claim that the automation of hardware-aware compression is essential to meet changes in the application and hardware architecture. To prove this claim and show that our proposed method is hardware-agnostic, we have applied our search method to two different hardware architectures, SiLago and Bitfusion. Those architectures have been selected due to the varying precision they support. In Sections~\ref{sec:silago-exp} and \ref{sec:bitfusion-exp} we have shown two different sets of solutions. The findings show that compression can be done in different ways depending on the target platform.  
The differences between the speedup values that appear in Tables~\ref{tab:silago} and~\ref{tab:bitfusion} for SiLago and Bitfusion might imply that Bitfusion is faster than Silago. 
This difference in speed is not what this work is particularly investigating, since we compare the optimized solutions on a given architecture to the baseline running on the same architecture.

For our method to be entirely generic, it needs to support variations in the NN model, compression method, and hardware platform. In this work, we have applied the method on two different architectures: CGRA (SiLago) and systolic array architecture (Bitfusion). Concerning the variations in the NN model, we have applied the inference-only search experiments on one model. Post-training quantization has been applied successfully on several models in the literature. And since our inference-only search method mainly relies on the success of post-training quantization, we believe our method is generic enough to be applied to many NN models. However, the beacon-based search needs to be applied to more models of several depths to investigate if the used beacons were enough. Thus, we applied the beacon-based search on two models of different depths. The aspect that needs more work is the variation in the compression method. The possibility of applying the post-training version of the other compression methods should be investigated. Also, the idea of the beacon-based search needs to be examined on other compression techniques to see if it can be directly applied, modified, or replaced by another method that satisfies the need for considering retraining effect on different compression configurations within a reasonable time.

\label{sec:limit}

\section{Conclusion and Future Work}
To run Neural Network (NN) applications on edge devices, they need to fit into such devices' restricted memory and computation capabilities. Thus, the compression of NN models is vital before they can be deployed. 
One powerful compression method is quantization, where fewer bits are used for NN weights and activations.
To reinforce the quantization benefits, the selection of precisions for different layers weights and activation should be optimized by involving the hardware model and application constraints. 
Therefore we introduce MOHAQ, the Multi-objective Hardware-aware Quantization method that can automate the optimal per-layer bit-selection for different platforms and application constraints. Both hardware efficiency and the error rate are considered objectives during the optimization, and the designer can select between varying Pareto alternatives. We exemplify and evaluate MOHAQ on Simple Recurrent Unit (SRU)-based RNN models for speech recognition. 

We used post-training quantization as a quantization method to evaluate solutions in a wide search space within feasible time. Solutions represent the precision of different layers, weights and activations. However, post-training quantization fails to find solutions with acceptable accuracy values in some scenarios.
Therefore, we introduce a novel "beacon-based search" technique, where we retrain the pretrained model for a few selected solutions to create beacons. These beacons are then used as a basis for the post-training quantization for other solutions during the search instead of retraining all evaluated solutions.

We have shown that the SRU-based model can be compressed using post-training quantization up to 8x without any error increase and to 12x with a 1.2 percentage point (p.p.) increase in the error. Also, we have applied the multi-objective search to quantize the SRU model to run on two architectures, SiLago and Bitfusion. We found a different solution set for each platform to meet its specific constraints. 

On SiLago, using inference-only search, we have found a set of solutions that, with increases in the error rate to 0.3 p.p. can achieve 97\% of the maximum possible speedup together with an 86\% energy saving.  
On Bitfusion, we have shown the search results using both inference-only search and beacon-based search on two SRU-based models. Our Beacon-based search allows us to take retraining into account, and we, therefore, can find better-performing solutions combined with lower error rates. The beacon-based search decreased the error of high speed up solutions by up to 4.9 p.p. for one model and 3.7 p.p. for another model compared to inference-only search.



%

\ifCLASSOPTIONcompsoc
  \section*{Acknowledgments}
\else
  \section*{Acknowledgment}
\fi

This research is part
of the CERES research program funded by the ELLIIT strategic research initiative funded by the Swedish government and Vinnova FFI project SHARPEN, under grant agreement no. 2018-05001.

The authors would also like to acknowledge the contribution of Tiago Fernandes Cortinhal in setting up the Python libraries and Yu Yang in the thoughtful discussions about SiLago architecture.

\ifCLASSOPTIONcaptionsoff
  \newpage
\fi



%
\bibliographystyle{IEEEtran} 
\markright{References}

\begin{thebibliography}{10}
\providecommand{\url}[1]{#1}
\csname url@samestyle\endcsname
\providecommand{\newblock}{\relax}
\providecommand{\bibinfo}[2]{#2}
\providecommand{\BIBentrySTDinterwordspacing}{\spaceskip=0pt\relax}
\providecommand{\BIBentryALTinterwordstretchfactor}{4}
\providecommand{\BIBentryALTinterwordspacing}{\spaceskip=\fontdimen2\font plus
\BIBentryALTinterwordstretchfactor\fontdimen3\font minus
  \fontdimen4\font\relax}
\providecommand{\BIBforeignlanguage}[2]{{%
\expandafter\ifx\csname l@#1\endcsname\relax
\typeout{** WARNING: IEEEtran.bst: No hyphenation pattern has been}%
\typeout{** loaded for the language `#1'. Using the pattern for}%
\typeout{** the default language instead.}%
\else
\language=\csname l@#1\endcsname
\fi
#2}}
\providecommand{\BIBdecl}{\relax}
\BIBdecl

\bibitem{rezk2019recurrent}
N.~M. {Rezk} \emph{et~al.}, ``Recurrent neural networks: An embedded computing
  perspective,'' \emph{IEEE Access}, vol.~8, pp. 57\,967--57\,996, 2020.

\bibitem{Pruning-Dai-2017}
X.~Dai, H.~Yin, and N.~K. Jha, ``Nest: A neural network synthesis tool based on
  a grow-and-prune paradigm,'' \emph{IEEE Transactions on Computers}, vol.~68,
  no.~10, pp. 1487--1497, 2019.

\bibitem{lowp-binary-connect}
M.~Courbariaux, Y.~Bengio, and J.-P. David, ``Binaryconnect: Training deep
  neural networks with binary weights during propagations,'' in \emph{Advances
  in neural information processing systems}, 2015, pp. 3123--3131.

\bibitem{HW-FPGA-FINN-L-Xilinx}
V.~Rybalkin, A.~Pappalardo, M.~M. Ghaffar, G.~Gambardella, N.~Wehn, and
  M.~Blott, ``Finn-l: Library extensions and design trade-off analysis for
  variable precision lstm networks on fpgas,'' in \emph{2018 28th international
  conference on field programmable logic and applications (FPL)}.\hskip 1em
  plus 0.5em minus 0.4em\relax IEEE, 2018, pp. 89--897.

\bibitem{energyawarepruning-Yang-2016}
T.-J. Yang, Y.-H. Chen, and V.~Sze, ``Designing energy-efficient convolutional
  neural networks using energy-aware pruning,'' in \emph{Proceedings of the
  IEEE Conference on Computer Vision and Pattern Recognition}, 2017, pp.
  5687--5695.

\bibitem{postq-nahshan19}
Y.~Nahshan, B.~Chmiel, C.~Baskin, E.~Zheltonozhskii, R.~Banner, A.~M.
  Bronstein, and A.~Mendelson, ``Loss aware post-training quantization,''
  \emph{Machine Learning}, vol. 110, no.~11, pp. 3245--3262, 2021.

\bibitem{Netadapt}
T.-J. Yang, A.~Howard, B.~Chen, X.~Zhang, A.~Go, M.~Sandler, V.~Sze, and
  H.~Adam, ``Netadapt: Platform-aware neural network adaptation for mobile
  applications,'' in \emph{Proceedings of the European Conference on Computer
  Vision (ECCV)}, 2018, pp. 285--300.

\bibitem{deepiot-Yao-2017}
S.~Yao \emph{et~al.}, ``Compressing deep neural network structures for sensing
  systems with a compressor-critic framework,'' \emph{CoRR}, vol.
  abs/1706.01215, 2017.

\bibitem{MOOP-savic2002}
D.~Savic, ``Single-objective vs. multiobjective optimisation for integrated
  decision support,'' \emph{Proceedings of the First Biennial Meeting of the
  International Environmental Modelling and Software Society}, vol.~1, pp.
  7--12, 01 2002.

\bibitem{postq-ocs-zhao19}
R.~Zhao, Y.~Hu, J.~Dotzel, C.~De~Sa, and Z.~Zhang, ``Improving neural network
  quantization without retraining using outlier channel splitting,'' in
  \emph{International conference on machine learning}.\hskip 1em plus 0.5em
  minus 0.4em\relax PMLR, 2019, pp. 7543--7552.

\bibitem{postq-banner19}
R.~Banner, Y.~Nahshan, and D.~Soudry, ``Post training 4-bit quantization of
  convolutional networks for rapid-deployment,'' in \emph{Advances in Neural
  Information Processing Systems 32}, H.~Wallach, H.~Larochelle,
  A.~Beygelzimer, F.~d\textquotesingle Alch\'{e}-Buc, E.~Fox, and R.~Garnett,
  Eds.\hskip 1em plus 0.5em minus 0.4em\relax Curran Associates, Inc., 2019,
  pp. 7950--7958.

\bibitem{postq-nagel20}
M.~Nagel, R.~A. Amjad, M.~Van~Baalen, C.~Louizos, and T.~Blankevoort, ``Up or
  down? adaptive rounding for post-training quantization,'' in
  \emph{International Conference on Machine Learning}.\hskip 1em plus 0.5em
  minus 0.4em\relax PMLR, 2020, pp. 7197--7206.

\bibitem{RNN-LSTM}
S.~Hochreiter and J.~Schmidhuber, ``Long short-term memory,'' \emph{Neural
  Comput.}, vol.~9, no.~8, pp. 1735--1780, Nov. 1997.

\bibitem{RNN-SRU}
T.~Lei, Y.~Zhang, and Y.~Artzi, ``Training {RNNs} as fast as {CNNs},''
  \emph{CoRR}, vol. abs/1709.02755, 2017.

\bibitem{pytorchkaldi-ravanelli2019}
M.~Ravanelli, T.~Parcollet, and Y.~Bengio, ``The pytorch-kaldi speech
  recognition toolkit,'' in \emph{ICASSP 2019-2019 IEEE International
  Conference on Acoustics, Speech and Signal Processing (ICASSP)}.\hskip 1em
  plus 0.5em minus 0.4em\relax IEEE, 2019, pp. 6465--6469.

\bibitem{silago-Hemani-2017}
A.~Hemani \emph{et~al.}, \emph{The SiLago Solution: Architecture and Design
  Methods for a Heterogeneous Dark Silicon Aware Coarse Grain Reconfigurable
  Fabric}.\hskip 1em plus 0.5em minus 0.4em\relax Cham: Springer International
  Publishing, 2017, pp. 47--94.

\bibitem{bitfusion-sharma-2018}
H.~Sharma, J.~Park, N.~Suda, L.~Lai, B.~Chau, V.~Chandra, and H.~Esmaeilzadeh,
  ``Bit fusion: Bit-level dynamically composable architecture for accelerating
  deep neural network,'' in \emph{2018 ACM/IEEE 45th Annual International
  Symposium on Computer Architecture (ISCA)}.\hskip 1em plus 0.5em minus
  0.4em\relax IEEE, 2018, pp. 764--775.

\bibitem{rnn-bi-lstm}
J.~Li and Y.~Shen, ``Image describing based on bidirectional {LSTM} and
  improved sequence sampling,'' in \emph{2017 IEEE 2nd International Conference
  on Big Data Analysis {(ICBDA)}}, March 2017, pp. 735--739.

\bibitem{RNN-projection}
H.~Sak, A.~Senior, and F.~Beaufays, ``Long short-term memory recurrent neural
  network architectures for large scale acoustic modeling,'' in \emph{Fifteenth
  annual conference of the international speech communication association},
  2014.

\bibitem{HW-FPGA-CLSTM}
S.~Wang, Z.~Li, C.~Ding, B.~Yuan, Q.~Qiu, Y.~Wang, and Y.~Liang, ``{C-LSTM:}
  enabling efficient {LSTM} using structured compression techniques on
  {FPGAs},'' in \emph{Proceedings of the 2018 ACM/SIGDA International Symposium
  on Field-Programmable Gate Arrays}, 2018, pp. 11--20.

\bibitem{post-quantize-samsung-Fang}
J.~Fang, A.~Shafiee, H.~Abdel-Aziz, D.~Thorsley, G.~Georgiadis, and J.~H.
  Hassoun, ``Post-training piecewise linear quantization for deep neural
  networks,'' in \emph{European Conference on Computer Vision}.\hskip 1em plus
  0.5em minus 0.4em\relax Springer, 2020, pp. 69--86.

\bibitem{UNPU-2019}
J.~Lee \emph{et~al.}, ``Unpu: An energy-efficient deep neural network
  accelerator with fully variable weight bit precision,'' \emph{IEEE Journal of
  Solid-State Circuits}, vol.~54, no.~1, pp. 173--185, 2019.

\bibitem{Sung_MMSE_2015}
W.~Sung, S.~Shin, and K.~Hwang, ``Resiliency of deep neural networks under
  quantization,'' \emph{CoRR}, vol. abs/1511.06488, 2015.

\bibitem{optimization_chong}
E.~K.~P. Chong and S.~H. Żak, \emph{An Introduction to Optimization, Fourth
  Edition}.\hskip 1em plus 0.5em minus 0.4em\relax John Wiley and Sons, Ltd,
  2013.

\bibitem{review-GA-chahar}
V.~Chahar, S.~Katoch, and S.~Chauhan, ``A review on genetic algorithm: Past,
  present, and future,'' \emph{Multimedia Tools and Applications}, vol.~80, 02
  2021.

\bibitem{NPGA-Horn-1994}
J.~Horn, N.~Nafpliotis, and D.~Goldberg, ``A niched pareto genetic algorithm
  for multiobjective optimization,'' in \emph{Proceedings of the First IEEE
  Conference on Evolutionary Computation. IEEE World Congress on Computational
  Intelligence}, 1994, pp. 82--87 vol.1.

\bibitem{NSGA-Srinivas-2000}
N.~Srinivas and K.~Deb, ``Muiltiobjective optimization using nondominated
  sorting in genetic algorithms,'' \emph{Evolutionary Computation}, vol.~2,
  no.~3, pp. 221--248, 1994.

\bibitem{NSGA2-Deb-2002}
K.~Deb, A.~Pratap, S.~Agarwal, and T.~Meyarivan, ``A fast and elitist
  multiobjective genetic algorithm: Nsga-ii,'' \emph{IEEE Transactions on
  Evolutionary Computation}, vol.~6, no.~2, pp. 182--197, 2002.

\bibitem{NSGA2-overview-2011}
Y.~Yusoff, M.~S. Ngadiman, and A.~M. Zain, ``Overview of nsga-ii for optimizing
  machining process parameters,'' \emph{Procedia Engineering}, vol.~15, pp.
  3978--3983, 2011, cEIS 2011.

\bibitem{sliding09}
M.~A. Shami and A.~Hemani, ``{Partially reconfigurable interconnection network
  for dynamically reprogrammable resource array},'' in \emph{International
  Conference on ASIC}.\hskip 1em plus 0.5em minus 0.4em\relax IEEE, oct 2009,
  pp. 122--125.

\bibitem{NACU20}
G.~Baccelli \emph{et~al.}, ``{NACU: A Non-Linear Arithmetic Unit for Neural
  Networks},'' in \emph{Design Automation Conference (DAC)}, vol.
  2020-July.\hskip 1em plus 0.5em minus 0.4em\relax IEEE, jul 2020, pp. 1--6.

\bibitem{vedic08}
H.~D. Tiwari \emph{et~al.}, ``{Multiplier design based on ancient Indian Vedic
  Mathematics},'' in \emph{International SoC Design Conference}, vol.~2.\hskip
  1em plus 0.5em minus 0.4em\relax IEEE, nov 2008, pp. II--65--II--68.

\bibitem{DiMArch16}
M.~A. Tajammul \emph{et~al.}, ``{TransMem: A memory architecture to support
  dynamic remapping and parallelism in low power high performance CGRAs},'' in
  \emph{International Workshop on Power and Timing Modeling, Optimization and
  Simulation (PATMOS)}.\hskip 1em plus 0.5em minus 0.4em\relax IEEE, 2016, pp.
  92--99.

\bibitem{sru-opt-shangguan19}
Y.~Shangguan \emph{et~al.}, ``Optimizing speech recognition for the edge,''
  \emph{CoRR}, vol. abs/1909.12408, 2019.

\bibitem{energyconstrained-Yang-2019}
H.~Yang, Y.~Zhu, and J.~Liu, ``End-to-end learning of energy-constrained deep
  neural networks,'' \emph{CoRR}, vol. abs/1806.04321, 2018.

\bibitem{timeconstr-Rizakis-2018}
M.~Rizakis, S.~I. Venieris, A.~Kouris, and C.-S. Bouganis, ``Approximate
  fpga-based lstms under computation time constraints,'' in \emph{International
  Symposium on Applied Reconfigurable Computing}.\hskip 1em plus 0.5em minus
  0.4em\relax Springer, 2018, pp. 3--15.

\bibitem{HAQ_Wang2019}
K.~Wang \emph{et~al.}, ``Haq: Hardware-aware automated quantization with mixed
  precision,'' in \emph{Proceedings of the IEEE/CVF Conference on Computer
  Vision and Pattern Recognition (CVPR)}, June 2019.

\bibitem{postq-zeroq-cai20}
Y.~Cai, Z.~Yao, Z.~Dong, A.~Gholami, M.~W. Mahoney, and K.~Keutzer, ``Zeroq: A
  novel zero shot quantization framework,'' in \emph{Proceedings of the
  IEEE/CVF Conference on Computer Vision and Pattern Recognition}, 2020, pp.
  13\,169--13\,178.

\bibitem{pymoo}
J.~{Blank} and K.~{Deb}, ``Pymoo: Multi-objective optimization in python,''
  \emph{IEEE Access}, vol.~8, pp. 89\,497--89\,509, 2020.

\bibitem{Rezk-distGA-2014}
N.~M. Rezk \emph{et~al.}, ``A distributed genetic algorithm for swarm robots
  obstacle avoidance,'' in \emph{2014 9th International Conference on Computer
  Engineering Systems (ICCES)}, 2014, pp. 170--174.

\bibitem{swarm-beacons-2013}
Y.~Tan and Z.~yang Zheng, ``Research advance in swarm robotics,'' \emph{Defence
  Technology}, vol.~9, no.~1, pp. 18--39, 2013.

\bibitem{dataset-timit}
J.~Garofolo and others., ``{DARPA TIMIT} acoustic-phonetic continous speech
  corpus cd-rom. nist speech disc 1-1.1,'' \emph{NASA STI/Recon Technical
  Report N}, vol.~93, p. 27403, 01 1993.

\bibitem{kaldi_ASRU2011}
D.~Povey \emph{et~al.}, ``The kaldi speech recognition toolkit,'' in \emph{IEEE
  2011 Workshop on Automatic Speech Recognition and Understanding}.\hskip 1em
  plus 0.5em minus 0.4em\relax IEEE Signal Processing Society, Dec. 2011, iEEE
  Catalog No.: CFP11SRW-USB.

\bibitem{NMTquant-Aji2019}
A.~F. Aji and K.~Heafield, ``Neural machine translation with 4-bit precision
  and beyond,'' \emph{CoRR}, vol. abs/1909.06091, 2019.

\end{thebibliography}

%






\end{document}